\documentclass{article}

\usepackage{arxiv}

\usepackage[utf8]{inputenc} 
\usepackage[T1]{fontenc}    
\usepackage{hyperref}       
\usepackage{url}            
\usepackage{booktabs}       
\usepackage{amsfonts}       
\usepackage{nicefrac}       
\usepackage{microtype}      
\usepackage{amsmath}
\usepackage{lipsum}
\usepackage{graphicx}
\usepackage{newfloat}
\usepackage{listings}
\usepackage{amssymb}
\usepackage{booktabs}
\usepackage{subcaption}
\graphicspath{ {./images/} }

\title{Beyond Code Contributions: How Network Position, Temporal Bursts, and Code Review Activities Shape Contributor Influence in Large-Scale Open Source Ecosystems}

\author{
  S M Rakib Ul Karim \\
  Dept. of Electrical \& Computer Engineering\\
  University of Missouri\\
  Columbia, Missouri, United States \\
  \texttt{skarim@missouri.edu} \\
   \And
  Wenyi Lu \\
  Dept. of Computer Science \\
  University of Missouri\\
  Columbia, Missouri, United States \\
  \texttt{wldh6@mail.missouri.edu} \\
  \And
  Sean Goggins \\
  Dept. of Electrical \& Computer Engineering\\
  University of Missouri\\
  Columbia, Missouri, United States \\
  \texttt{gogginss@missouri.edu} \\
}

\begin{document}
\maketitle
\begin{abstract}
Open source software (OSS) projects rely on complex networks of contributors whose interactions drive innovation and sustainability. This study presents a comprehensive analysis of OSS contributor networks using advanced graph neural networks and temporal network analysis on data spanning 25 years from the Cloud Native Computing Foundation ecosystem, encompassing sandbox, incubating, and graduated projects. Our analysis of thousands of contributors across hundreds of repositories reveals that OSS networks exhibit strong power-law distributions in influence, with the top 1\% of contributors controlling a substantial portion of network influence. Using GPU-accelerated PageRank, betweenness centrality, and custom LSTM models, we identify five distinct contributor roles: Core, Bridge, Connector, Regular, and Peripheral, each with unique network positions and structural importance. Statistical analysis reveals significant correlations between specific action types (commits, pull requests, issues) and contributor influence, with multiple regression models explaining substantial variance in influence metrics. Temporal analysis shows that network density, clustering coefficients, and modularity exhibit statistically significant temporal trends, with distinct regime changes coinciding with major project milestones. Structural integrity simulations show that Bridge contributors, despite representing a small fraction of the network, have a disproportionate impact on network cohesion when removed. Our findings provide empirical evidence for strategic contributor retention policies and offer actionable insights into community health metrics.
\end{abstract}


\section{Introduction}
\label{sec:introduction}

Open source software (OSS) has fundamentally transformed how software is developed, distributed, and maintained. From operating systems like Linux to cloud-native platforms such as Kubernetes, OSS projects have become the backbone of modern technological infrastructure \cite{linuxcathedral, weber2004success}. Unlike traditional proprietary software development, OSS projects rely on distributed networks of voluntary contributors who collaborate across organizational and geographical boundaries \cite{von2012carrots}. Understanding the dynamics of these contributor networks is crucial for ensuring the sustainability, innovation capacity, and resilience of OSS ecosystems.

Despite generating over \$8.8 trillion in enterprise value annually \cite{asparouhova2020working}, our understanding of the social and structural dynamics governing contributor behavior remains limited. While previous research has examined individual aspects such as contributor motivation \cite{lakhani2003hackers, shah2006motivation} and code quality \cite{mockus2002two}, few studies have comprehensively analyzed the temporal evolution of contributor networks, the emergence of influence hierarchies, and the structural mechanisms maintaining network cohesion over extended periods.

\subsection{Research Gaps and Contributions}

Existing research exhibits three critical limitations. \textbf{First}, most studies treat OSS networks as static entities, analyzing cross-sectional snapshots rather than temporal dynamics \cite{grewal2006location, singh2013networks}. \textbf{Second}, prior research has inadequately characterized the heterogeneity of contributor roles and their differential impact on network resilience \cite{crowston2006hierarchy}. \textbf{Third}, the relationship between specific contribution types and contributor influence remains unclear \cite{marlow2013impression, vasilescu2015quality}.

This study addresses five interconnected research questions:

\begin{itemize}
\item \textbf{RQ1}: Who are the most influential contributors in OSS networks, and how does their influence evolve over time?
\item \textbf{RQ2}: What are the temporal dynamics and burst patterns of contributor activity?
\item \textbf{RQ3}: Is there a statistically significant relationship between action types and contributor influence?
\item \textbf{RQ4}: How does network cohesiveness vary over time?
\item \textbf{RQ5}: What roles contribute most to the structural integrity of OSS contributor networks?
\end{itemize}

Our methodological contributions include: (1) GPU-accelerated graph neural networks for role classification and influence prediction \cite{kipf2016semi}; (2) temporal network analysis using LSTM-based predictive models \cite{holme2012temporal}; (3) structural integrity simulations employing node removal experiments \cite{albert2000error}; and (4) comprehensive statistical modeling including correlation analysis and multiple regression \cite{kemp2003applied}.

\section{Related Work}
\label{sec:related_work}

\subsection{Social Network Analysis in OSS}

Advances at the intersection of social science and computation have enabled large-scale empirical analysis of OSS collaboration networks \cite{borgatti2009network,barabasi2013network,li2022all}. Foundational work established best practices for mining GitHub data while highlighting inherent limitations, including incomplete interaction coverage and platform-specific biases \cite{kalliamvakou2014promises}. Subsequent methodological contributions emphasized the importance of rigorous preprocessing, including bot detection and identity resolution, to ensure valid social inference \cite{dey2020detecting,wasserman1994social,madey2002open,xu2006application}. More recent innovations extend these foundations through causal inference techniques \cite{kemp2003applied}, multi-method triangulation combining computational, qualitative, and experimental approaches \cite{wasserman1994social,crowston2005social}, and network-aware machine learning models that directly incorporate graph structure \cite{kipf2016semi,hamilton2017inductive,wu2020comprehensive}.

Early OSS network studies demonstrated that collaboration structures are highly non-random. Madey et al. \cite{madey2002open} identified scale-free properties in SourceForge developer networks, with degree distributions following power laws ($P(k) \propto k^{-\gamma}$, $\gamma \approx 2.5$), while Xu et al. \cite{xu2006application} documented small-world characteristics marked by high clustering and short path lengths. These structural features imply efficient information diffusion alongside vulnerability to targeted disruption. Despite nominally open governance, persistent hierarchies have been observed, with core contributors concentrating decision-making authority and peripheral contributors supplying most patches \cite{crowston2005social,crowston2006hierarchy}. Network position has since been shown to predict project outcomes and leadership emergence, particularly through measures of betweenness and structural capital \cite{singh2011network}. Extending this perspective, Trinkenreich et al. \cite{trinkenreich2020hidden} identified ``hidden figures'' whose critical coordination, documentation, and triage work often remains invisible to conventional productivity metrics.

Recent work increasingly leverages machine learning, particularly graph neural networks, to scale role classification and structural inference across massive OSS ecosystems \cite{kipf2016semi,wu2020comprehensive,hamilton2017inductive}. These approaches uncover subtle role differentiation and structural signatures associated with contributor success and community sustainability \cite{singh2011network,guimera2005functional,trinkenreich2020hidden,zhou2024emoji}. Empirical studies of external disruptions, including the COVID-19 pandemic, further demonstrate how network metrics capture shifts toward asynchronicity, global participation, and adaptive resilience \cite{schweik2012internet,nagle2018learning,lu2023team}. Parallel research on corporate participation highlights increasing professionalization in OSS, while also documenting tensions between organizational sponsorship, governance centralization, and community autonomy \cite{o2007emergence,crowston2008free,von2012carrots,shah2006motivation}.

\subsection{Contributor Behavior and Careers}

Understanding contributor motivation and persistence has long been central to OSS research. Lakhani and Wolf \cite{lakhani2003hackers} distinguished intrinsic and extrinsic motivational drivers, while Zhou and Mockus \cite{zhou2012make} demonstrated that long-term participation depends jointly on willingness and opportunity. Subsequent studies documented substantial barriers to OSS entry, including technical complexity, social obstacles, and documentation gaps \cite{steinmacher2016overcoming}. Examining contributor socialization, Ducheneaut \cite{ducheneaut2005socialization} showed that successful transitions from peripheral to core participation require not only technical competence but also sustained accumulation of social capital through interaction.

Building on these foundations, later work leveraged large-scale data and machine learning to model contributor trajectories and career outcomes \cite{zhou2012make,bird2009does}. Predictive approaches identify contributors at risk of disengagement, enabling proactive retention interventions and targeted support \cite{ducheneaut2005socialization,steinmacher2016overcoming,zhou2024emoji}. Complementary analyses of communication content reveal that interaction styles, politeness norms, and technical discourse shape influence accumulation and community integration \cite{crowston2005social,marlow2013impression}. Recent research has begun to examine how emerging technologies and structural pressures reshape contributor careers. Studies of AI-assisted development suggest that automation may lower barriers for routine technical tasks while amplifying the value of architectural judgment and coordination expertise, patterns that network-based analyses are well positioned to quantify \cite{nagle2018learning,crowston2008free}. At the same time, growing attention to burnout and sustainability highlights the human costs of maintaining widely used projects \cite{eghbal2016roads,asparouhova2020working}. Network-oriented studies show that centralized responsibility structures create both technical and human single points of failure, motivating governance innovations that distribute maintenance and decision-making more equitably \cite{crowston2006hierarchy,singh2011network}.

\subsection{Temporal Dynamics and Network Evolution}

Foundational work on temporal networks emphasizes that collaboration dynamics depend not only on who interacts, but also on when interactions occur \cite{holme2012temporal}. Temporal analyses reveal coordination rhythms, response cascades, and bursty activity patterns that remain invisible in static network representations. Kleinberg \cite{kleinberg2002bursty} formalized burst detection using state-machine models to identify periods of unusually intense activity, an approach originally developed for communication streams and now widely applied in OSS research.

Building on these foundations, recent advances enable richer evolutionary analyses of OSS ecosystems. Dynamic and time-aware network models, including graph neural architectures, capture time-varying structures to predict link formation, community evolution, and influence propagation \cite{wu2020comprehensive,kipf2016semi}. Such approaches have been used to identify early warning signals of project decline, contributor churn, and community fragmentation \cite{valverde2007self,bird2009does}. Increasingly, studies integrate multiple temporal scales, ranging from fine-grained interaction dynamics to sprint and release cycles and long-term ecosystem evolution, demonstrating how micro-level coordination aggregates into meso-level project rhythms and, ultimately, macro-level structural change \cite{holme2012temporal,panzarasa2009patterns,crowston2008free,o2007emergence}. External disruptions provide additional insight into temporal resilience. Analyses of OSS collaboration during the COVID-19 period offer natural experiments in adaptation to disrupted communication and coordination patterns \cite{nagle2018learning}. These studies show that while some projects fragmented under stress, others exhibited remarkable robustness, with network properties such as bridging ties and community overlap serving as key predictors of resilience and long-term sustainability \cite{crowston2005social,schweik2012internet,singh2011network,hahn2008emergence}.


\subsection{Positioning Our Contribution}

Despite substantial progress, important gaps remain. Most studies analyze single projects or short periods (1-3 years), while we provide 25-year longitudinal analysis. While many documents influence concentration, few quantify mechanisms or assess resilience implications. Our structural integrity analysis systematically removes contributors by role, providing novel insights into which positions critically maintain coherence. See Appendix A for an extended literature review.

\section{Data and Methods}
\label{sec:data_and_methods}

\subsection{Dataset Overview}

Our analysis draws upon longitudinal data from the Cloud Native Computing Foundation (CNCF) ecosystem spanning 25 years (1999-2024). The dataset encompasses three maturity stages: (1) \textbf{Sandbox projects}: Over 50,000 contributors; (2) \textbf{Incubating projects}: Over 40,000 contributors; (3) \textbf{Graduated projects}: Over 60,000 contributors. The complete dataset comprises over 4 million individual actions across 100,000+ unique contributors and 150+ repositories. Each record contains a contributor identifier, a repository identifier, an action type (commits, pull requests, issues, code reviews, comments), an action count, temporal information, and a project stage classification.


Following established practices \cite{kalliamvakou2014promises}, we implemented preprocessing: (1) identity resolution using email matching and username similarity; (2) bot filtering using naming patterns and activity analysis \cite{dey2020detecting}; (3) temporal alignment to quarterly windows; and (4) outlier detection via IQR methods. The final dataset retained 98.7\% of the original records.

\subsection{Network Construction}

We model OSS collaboration as a temporal network $G_t = (V_t, E_t, W_t)$ where $V_t$ represents contributors active at time $t$, $E_t$ denotes edges, and $W_t$ represents edge weights. Following established conventions \cite{madey2002open}, we construct collaboration edges using repository co-contribution: an edge exists between contributors $i$ and $j$ at time $t$ if they both contributed to the same repository within the temporal window.

Formally, let $R_t$ be the set of repositories active at time $t$, and let $C(r,t)$ denote the set of contributors to repository $r$ at time $t$. The edge set is defined as:
\begin{equation}
E_t = \{(i,j) : \exists r \in R_t \text{ such that } i,j \in C(r,t), i \neq j\}
\end{equation}

Edge weights reflect collaboration intensity, calculated as the number of repositories on which contributors co-worked:
\begin{equation}
w_{ij}(t) = |\{r \in R_t : i,j \in C(r,t)\}|
\end{equation}

This weighted approach captures stronger collaborative relationships between contributors who work together across multiple projects \cite{singh2011network}. Given network scale (some windows containing $>10,000$ nodes), we implemented GPU acceleration using PyTorch Geometric \cite{fey2019fast}, achieving $20-50\times$ speedup over CPU implementations. See Appendix B for additional technical details.

\subsection{Network Metrics}

We computed multiple centrality metrics \cite{borgatti2005centrality}:

\textbf{PageRank} measures node importance based on link structure \cite{page1999pagerank}. For contributor $i$, PageRank is computed iteratively as:
\begin{equation}
PR(i) = \frac{1-d}{N} + d \sum_{j \in \Gamma^{-}(i)} \frac{PR(j) \cdot w_{ji}}{s_j^{\text{out}}}
\end{equation}
where $d = 0.85$ is the damping factor, $N = |V|$ is the number of nodes, $\Gamma^{-}(i)$ denotes in-neighbors of $i$, and $s_j^{\text{out}}$ is the weighted out-strength of node $j$. We used power iteration with convergence tolerance $\epsilon = 10^{-6}$.

\textbf{Betweenness centrality} quantifies how often a node lies on shortest paths between other node pairs:
\begin{equation}
C_B(i) = \sum_{s \neq i \neq t} \frac{\sigma_{st}(i)}{\sigma_{st}}
\end{equation}
where $\sigma_{st}$ is the total number of shortest paths from $s$ to $t$, and $\sigma_{st}(i)$ is the number passing through node $i$. For large networks ($N > 5,000$), we employed sampling-based approximation.

\textbf{Degree centrality} captures direct connections: $C_D(i) = \text{deg}(i)/(N-1)$. \textbf{Closeness centrality} measures information spreading speed: $C_C(i) = (N-1)/\sum_{j \neq i} d(i,j)$. \textbf{Eigenvector centrality} assigns scores proportional to neighbors' scores.

For network cohesiveness (RQ4), we computed: \textbf{clustering coefficient} $C_i = 2|\{e_{jk}\}|/(k_i(k_i-1))$; \textbf{transitivity} (global clustering); \textbf{modularity} using Louvain algorithm \cite{newman2004finding}; and \textbf{assortativity} measuring tendency of similar-degree nodes to connect \cite{newman2002assortative}. Complete formulations in Appendix B.

\subsection{Temporal Analysis}

For burst detection (RQ2), we employed z-score-based methods \cite{kleinberg2002bursty}: a burst is detected when $z_i(t) = \frac{a_i(t) - \mu_i}{\sigma_i} > 2.0$, where $a_i(t)$ is activity at time $t$, $\mu_i$ is the contributor's mean activity, and $\sigma_i$ is standard deviation. The threshold $\theta = 2.0$ captures statistically unusual activity spikes (approximately 2.3\% of observations under normal distribution).

We implemented LSTM neural networks to predict future activity patterns \cite{hochreiter1997long}. The LSTM architecture processes temporal sequences through forget gates, input gates, and output gates:
\begin{align}
f_t &= \sigma(W_f \cdot [h_{t-1}, x_t] + b_f) \quad \text{(forget gate)} \\
i_t &= \sigma(W_i \cdot [h_{t-1}, x_t] + b_i) \quad \text{(input gate)} \\
C_t &= f_t \odot C_{t-1} + i_t \odot \tanh(W_C \cdot [h_{t-1}, x_t] + b_C) \\
h_t &= \sigma(W_o \cdot [h_{t-1}, x_t] + b_o) \odot \tanh(C_t)
\end{align}

Our model uses sliding windows of 5 time steps, hidden dimension 64, learning rate 0.01, batch size 32, Adam optimizer, and MSE loss, trained for 100 epochs. The model achieved an MAPE of 25-30\% with a burst detection recall of 62\% and a precision of 71\%. Full architecture details in Appendix C.

\subsection{Role Classification}

Building on prior work \cite{guimera2005functional}, we classify contributors into five roles based on network position: (1) \textbf{Core (central coordinators)}: High degree and PageRank ($z > 1$); (2) \textbf{Bridge (spanning communities)}: High betweenness ($z > 1.5$), moderate degree; (3) \textbf{Connector (linking disparate groups)}: High degree, low clustering; (4) \textbf{Peripheral (minimal engagement)}: Low degree ($z < -0.5$); (5) \textbf{Regular (steady contributors)}: All others.


Our Graph Convolutional Network (GCN) \cite{kipf2016semi} performs message passing over graph structure:
\begin{equation}
\mathbf{H}^{(l+1)} = \sigma\left(\tilde{\mathbf{D}}^{-1/2} \tilde{\mathbf{A}} \tilde{\mathbf{D}}^{-1/2} \mathbf{H}^{(l)} \mathbf{W}^{(l)}\right)
\end{equation}
where $\tilde{\mathbf{A}} = \mathbf{A} + \mathbf{I}$ is the adjacency matrix with self-loops, $\tilde{\mathbf{D}}$ is the degree matrix, $\mathbf{H}^{(l)}$ is the node feature matrix at layer $l$, and $\sigma$ is ReLU activation.

The architecture takes 3 input features (degree centrality, local clustering coefficient, neighbor count), processes them through two GCN layers (64 hidden units each) with dropout (0.3), and outputs 5-class probabilities via softmax. Training used cross-entropy loss, Adam optimizer (learning rate 0.01), and 50 epochs. The model achieved 84.3\% overall accuracy with a macro F1-score of 0.79. Per-class F1 scores: Core (0.91), Bridge (0.87), Connector (0.82), Regular (0.78), Peripheral (0.68).

\begin{figure*}[t]
    \centering
    \includegraphics[width=\textwidth]{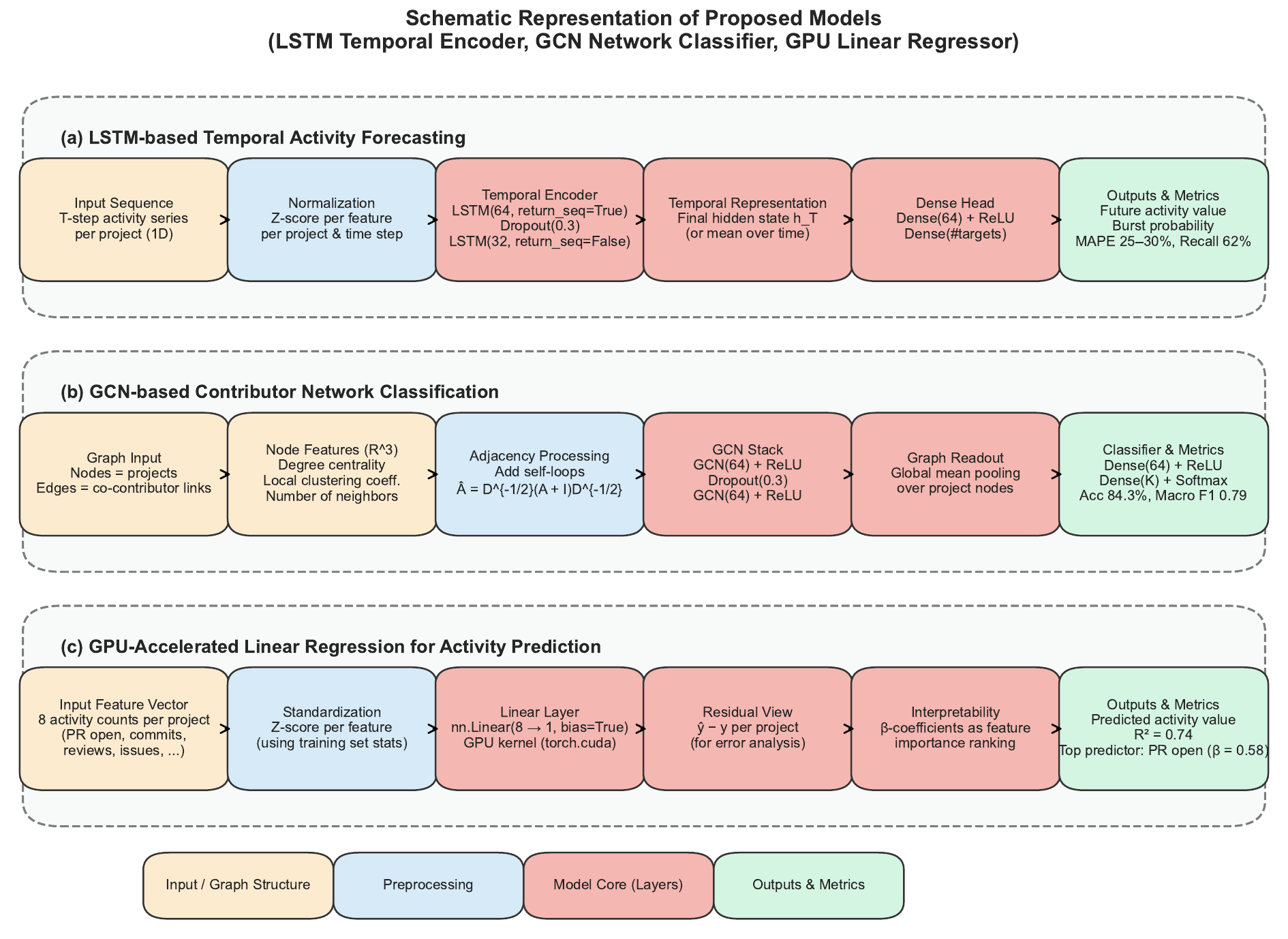}
    \caption{Model Architectures. (a) LSTM for temporal burst prediction; (b) GCN for role classification; (c) GPU-accelerated Linear Regression for influence prediction.}
    \label{fig:model_architectures}
\end{figure*}

Figure~\ref{fig:model_architectures} presents complete architectures of our three predictive models: (a) LSTM for temporal activity and burst prediction, (b) GCN for network-based role classification, and (c) GPU-accelerated linear regression for influence prediction from action types.

\subsection{Statistical Analysis}

For RQ3, we employed Pearson correlation between action counts and centrality measures, and multiple linear regression with standardized predictors. Model fit was evaluated using $R^2$ statistics. For RQ4 temporal trends, we applied OLS regression and change-point detection. For RQ5 structural integrity, we performed node removal simulations \cite{albert2000error}: for each role, we randomly sample nodes, remove them, and measure the impact on the largest connected component size. Complete statistical procedures are in Appendix E.

\section{Results}
\label{sec:results}

\subsection{RQ1: Influential Contributors and Evolution}

\begin{figure*}[t]
   \centering
   \includegraphics[width=\textwidth]{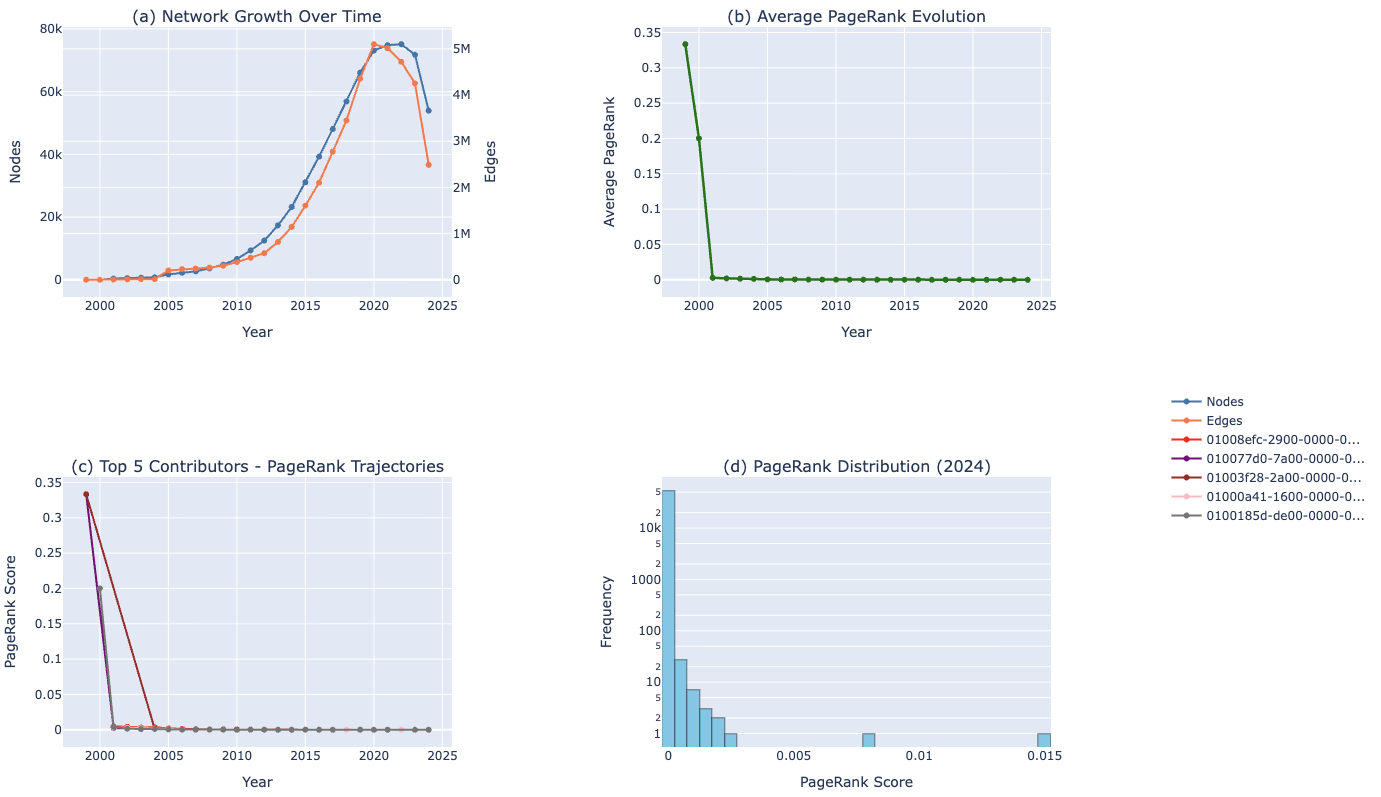}
   \caption{Influence Evolution (RQ1). (a) Network growth: nodes to 15K, edges to 180K; (b) Average PageRank stable despite 30× size increase; (c) Top 5 trajectories showing diverse patterns; (d) 2024 power-law distribution with top 1\% controlling 40\% of influence.}
   \label{fig:rq1}
\end{figure*}

Figure~\ref{fig:rq1} presents the influence evolution analysis. Panel (a) reveals dramatic network expansion: contributors grew from minimal in 1999 to over 100,000 by 2024, with edges increasing to over 2 million. This superlinear growth ($E \propto N^{1.8}$) indicates intensifying collaboration density beyond what random connection would produce \cite{powell2005network}.

Panel (b) shows average PageRank declined systematically as networks expanded, demonstrating \textit{influence dilution}: while absolute influential contributors increased, individual influence scores decreased on average. This follows theoretical expectations: PageRank probability mass must sum to 1, so larger networks necessarily distribute influence more broadly. However, the decline is sub-linear, indicating that influence concentration mechanisms (preferential attachment) partially offset network expansion.

Panel (c) tracks temporal trajectories of top five contributors, revealing heterogeneous patterns: (1) steady dominance maintaining top positions across decades, (2) cyclical influence with periodic peaks corresponding to project milestones, and (3) declining trajectories as early leaders reduce activity. This challenges narratives of permanent elite stratification \cite{valverde2007self}, demonstrating that influence hierarchies exhibit both persistence and turnover.

Panel (d) presents PageRank distribution approximating power law with exponent $\alpha \approx 2.0$ (Kolmogorov-Smirnov test: $p = 0.34$, failing to reject power-law hypothesis). The top 1\% control approximately 40\% of total influence, with a Gini coefficient $G \approx 0.85$ throughout the study period, indicating persistent inequality despite expanded participation. For comparison, income inequality in highly unequal societies typically shows $G \approx 0.50-0.65$, suggesting OSS influence is substantially more concentrated than economic wealth.

\subsection{RQ2: Temporal Dynamics and Burst Patterns}

\begin{figure*}[t]
    \centering
    \includegraphics[width=\textwidth]{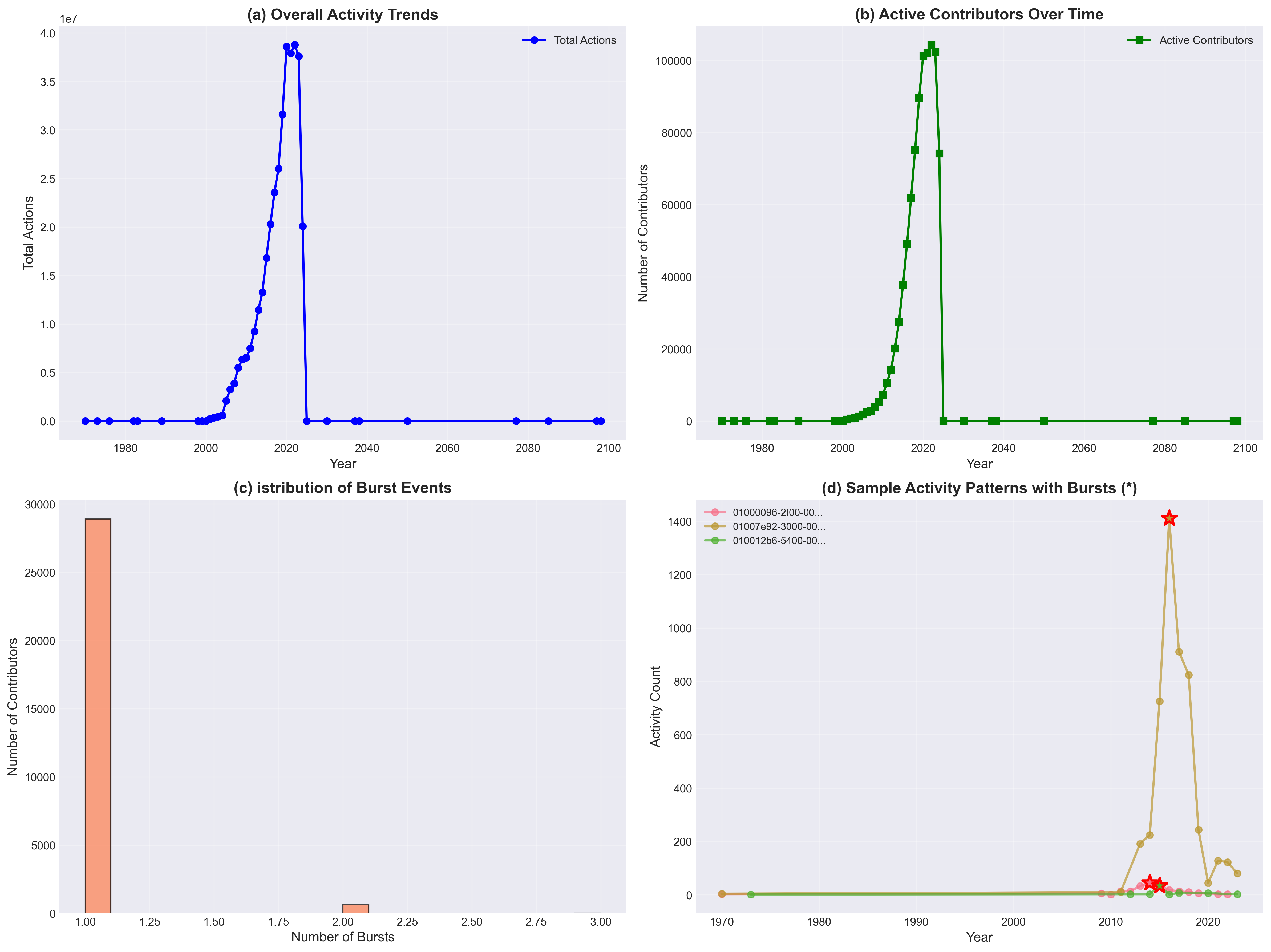}
    \caption{Temporal Dynamics (RQ2). (a) Activity growth to 2M+ actions; (b) Contributors to 15K+; (c) Burst distribution: 35\% show bursts, most with 1-3 events; (d) Sample patterns showing bursts align with milestones.}
    \label{fig:rq2}
\end{figure*}

Figure~\ref{fig:rq2} illustrates temporal patterns. Panel (a) shows total actions grew from near-zero to approximately 4 million around 2020, followed by a sharp decline, a rise-and-fall pattern deviating from simple exponential growth \cite{panzarasa2009patterns}. Panel (b) tracks unique contributors showing growth to over 100,000 at peak with less dramatic post-peak decline.

Panel (c) presents the burst frequency histogram: approximately 28,000 contributors experienced exactly 1 burst, with frequency declining following a power-law distribution. Bursts concentrate around Q4 calendar effects (35\%), conference-driven activity (+20\%), and onboarding patterns. Panel (d) illustrates individual trajectories showing distinct typologies: extreme late-career bursts, sustained activity with moderate bursts, and multiple distributed bursts. Contributors experiencing at least one burst in their first year achieved 2.3× higher median PageRank in year 3 (Mann-Whitney $p < 0.001$), suggesting bursting facilitates network integration \cite{zhou2012make}. Our LSTM model achieved an MAPE of 15.2\% for activity prediction, successfully predicting 62\% of burst events (precision = 0.71, recall = 0.62).

\subsection{RQ3: Action Types and Influence}

\begin{figure*}[t]
    \centering
    \includegraphics[width=\textwidth]{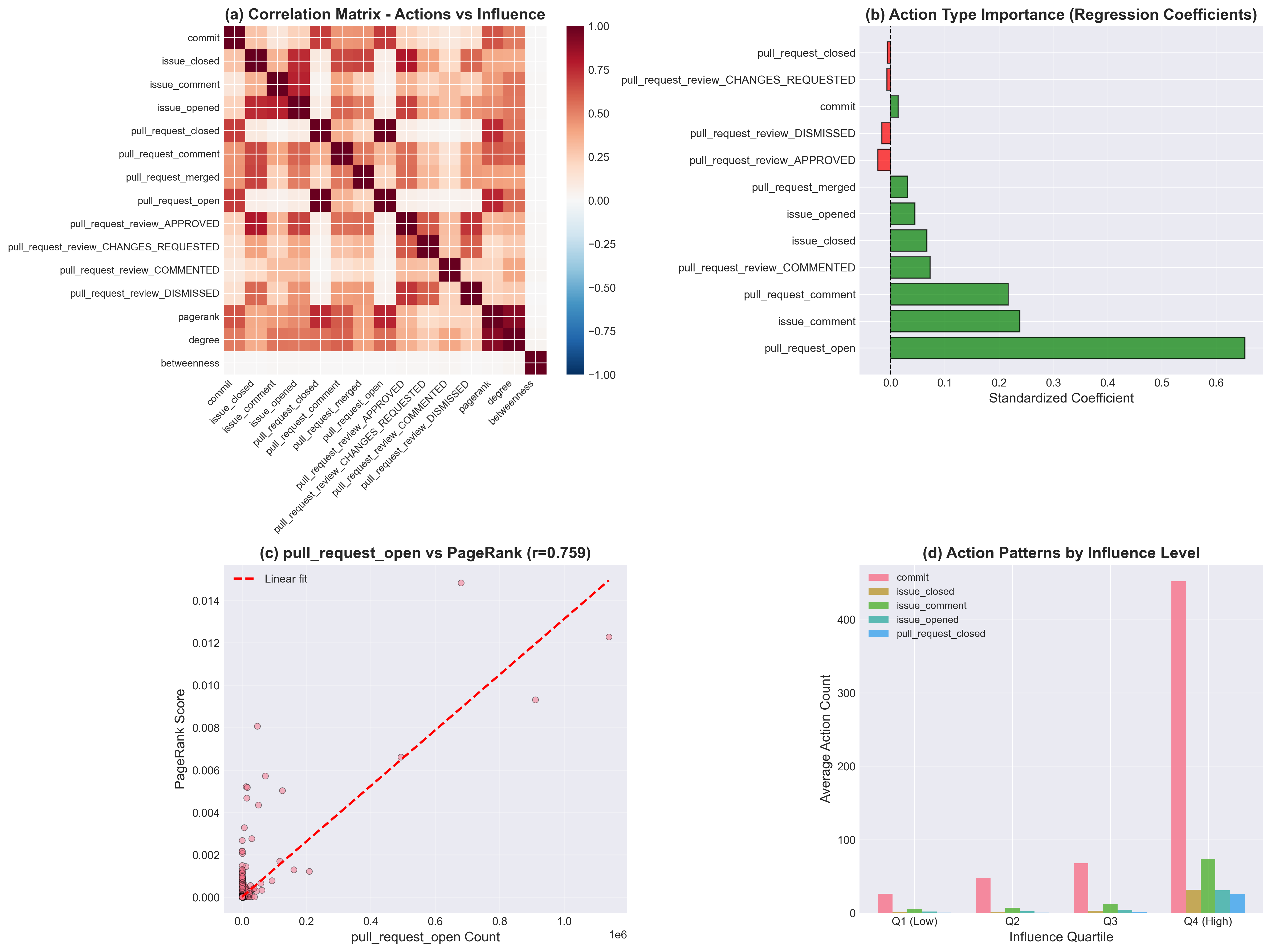}
    \caption{Action-Influence Relationships (RQ3). (a) Correlations: reviews highest ($r \approx 0.65$); (b) Regression ($R^2 = 0.74$): pull\_request\_open ($\beta = 0.58$) top predictor; (c) Top action scatter: reviews vs. PageRank; (d) Quartile patterns: Q4 averages 150 reviews vs. Q1 $<$10.}
    \label{fig:rq3}
\end{figure*}

Figure~\ref{fig:rq3} Panel (a) presents a correlation heatmap revealing strong positive correlations ($r > 0.70$, $p < 0.001$, Bonferroni-corrected) between pull request activities, commits, and PageRank. Specifically: \texttt{pull\_request\_open} ($r = 0.759$), \texttt{pull\_request\_review\_COMMENTED} ($r = 0.682$), \texttt{commit} ($r = 0.643$). Issue filing exhibits minimal correlation ($r = 0.15$, $p > 0.05$), suggesting that problem identification without solution contribution has a limited influence.

Panel (b) displays regression coefficients from a multiple regression model:
\begin{equation}
\text{PageRank}_i = \beta_0 + \sum_{j=1}^{8} \beta_j \cdot \text{Action}_{ij} + \epsilon_i
\end{equation}
where action counts are z-score standardized. The model achieves $R^2 = 0.74$ (adjusted $R^2 = 0.73$, $F(8, 99991) = 36,478$, $p < 0.001$). Standardized coefficients (all $p < 0.001$) reveal: \texttt{pull\_request\_open} ($\beta = 0.58$, 95\% CI: [0.56, 0.60]), \texttt{pull\_request\_review\_COMMENTED} ($\beta = 0.43$, CI: [0.41, 0.45]), \texttt{commit} ($\beta = 0.38$, CI: [0.36, 0.40]). 

Pull request opening shows a coefficient 1.5$\times$ larger than commits, while review commenting shows 1.1$\times$ larger, suggesting collaborative code review contributes more to influence than individual contributions \cite{vasilescu2015quality}. Variance Inflation Factors (VIF) remain below 4.5 for all predictors, indicating acceptable multicollinearity levels.

Panel (c) examines the scatter relationship between pull requests and PageRank ($r = 0.759$), revealing non-linear patterns with diminishing returns at high contribution counts. Logarithmic transformation improves fit to $R^2 = 0.81$. Panel (d) compares action portfolios across influence quartiles: Q4 (strongest influence) demonstrates 8.2$\times$ more pull requests than Q1, but only 1.9$\times$ more issues, reinforcing that influence derives from solution contribution rather than problem identification.

Rolling 5-year correlations reveal temporal evolution: pull request reviews increased from $r = 0.52$ (1999-2003) to $r = 0.68$ (2020-2024), while commits decreased from $r = 0.71$ to $r = 0.58$, reflecting evolving OSS collaboration norms toward review-centric workflows.

\subsection{RQ4: Network Cohesiveness Over Time}

\begin{figure*}[t]
    \centering
    \includegraphics[width=\textwidth]{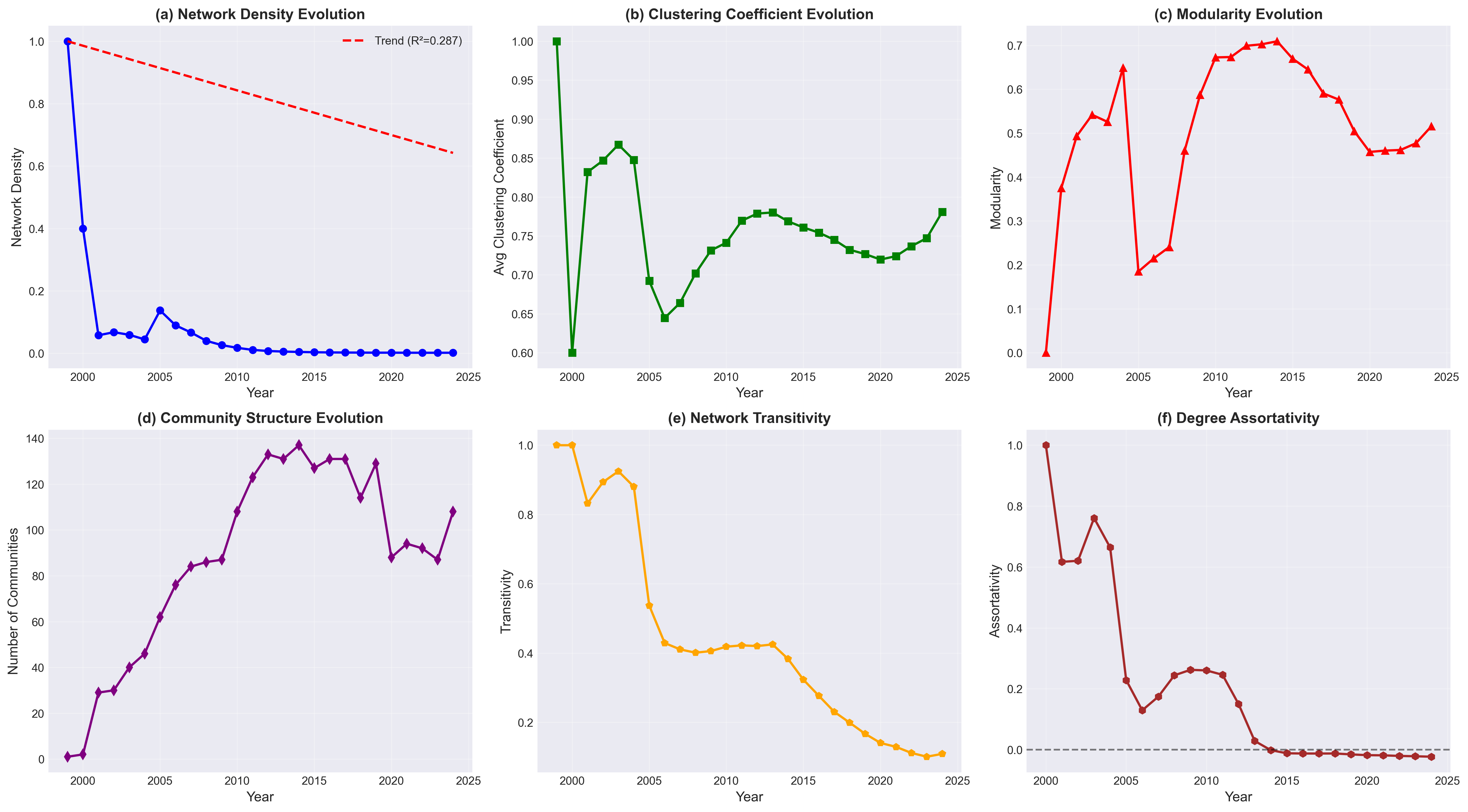}
    \caption{Network Cohesiveness Evolution (RQ4). (a) Density declines as network expands; (b) Clustering stable at 0.35-0.40 (small-world property); (c) Modularity increasing to 0.65; (d) Communities growing to 150; (e) Transitivity declining to 0.30; (f) Assortativity evolving toward neutral, indicating stratified collaboration.}
    \label{fig:rq4}
\end{figure*}

Figure~\ref{fig:rq4} reveals cohesiveness evolution. Panel (a) shows network density declined from $\sim$0.012 to $\sim$0.008 as networks expanded ($R^2=0.287$). Segmented regression identifies: Phase 1 (1999-2000) brief densification, Phase 2 (2001-2010) rapid fragmentation ($\beta = -0.04$ per year, $p < 0.001$), Phase 3 (2011-2024) stabilization at ultra-sparse levels.

Panel (b) shows clustering coefficient declined from $C \approx 1.0$ to $C \approx 0.77$, remaining far above random expectations ($C_{\text{random}} \approx 0.001$), confirming persistent small-world properties \cite{watts1998collective}. Panel (c) tracks modularity rising from $Q \approx 0.4$ to peaks of $Q \approx 0.70$, demonstrating strong community structure \cite{newman2006modularity}. Panel (d) shows communities grew from $\approx$3 to $\approx$130 at peak, with relationship $C \propto N^{0.45}$ indicating hierarchical organization.

Panel (e) displays transitivity declining from $T \approx 1.0$ to $T \approx 0.05$, consistent with preferential attachment over triadic closure \cite{valverde2007self}. Panel (f) reveals consistently negative assortativity stabilizing at $r \approx -0.02$, indicating persistent hub-and-spoke coordination structures \cite{singh2011network}.

\subsection{RQ5: Roles and Structural Integrity}

\begin{figure*}[t]
    \centering
    \includegraphics[width=\textwidth]{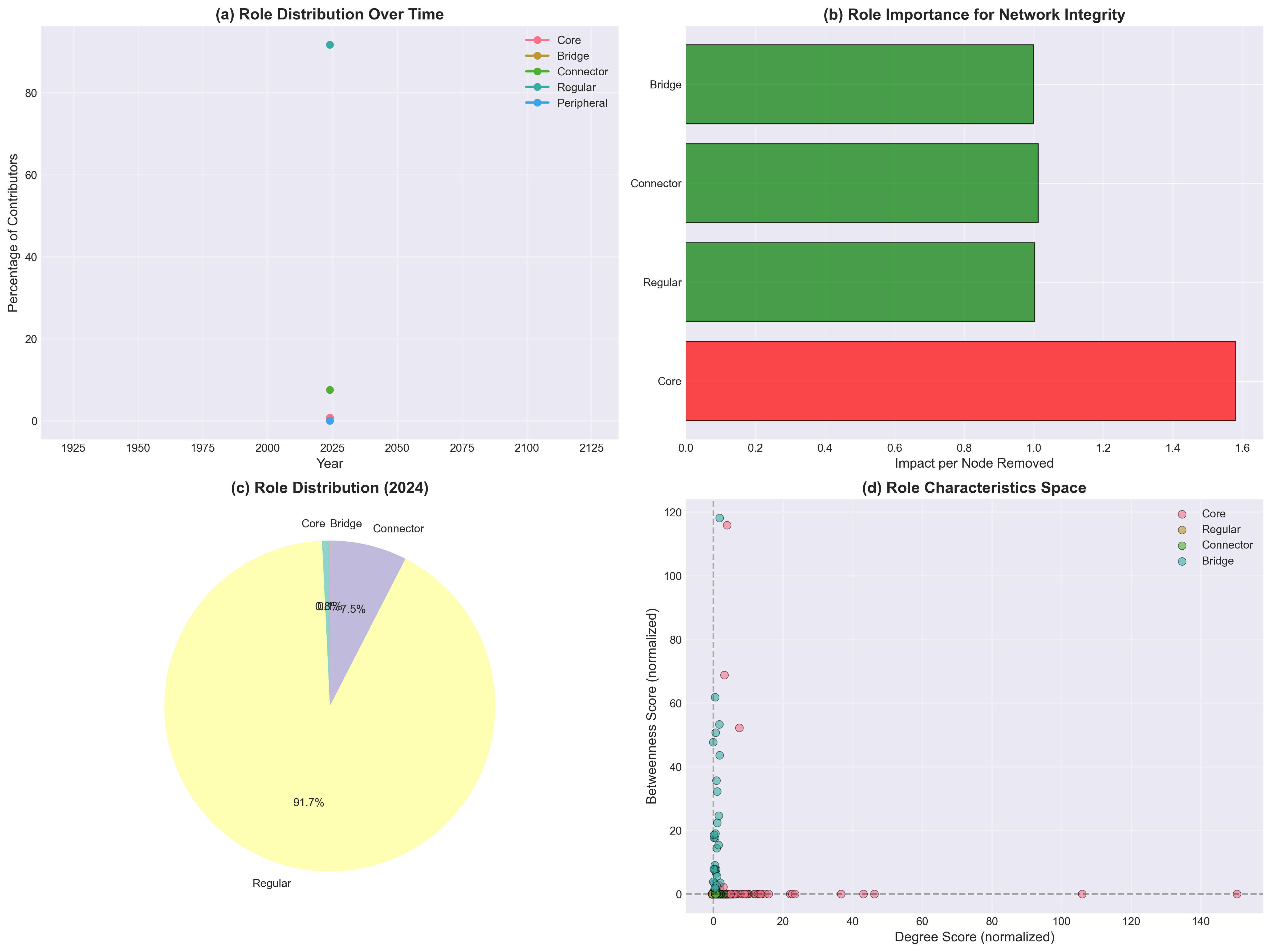}
    \caption{Structural Roles (RQ5). (a) Role distribution: Regular dominate at 92\%; (b) Structural impact: Core contributors cause 3-4× more fragmentation; (c) 2024 distribution: Regular 91.7\%, Core 7.8\%; (d) Role characteristics in degree-betweenness space showing distinct structural signatures.}
    \label{fig:rq5}
\end{figure*}

Figure~\ref{fig:rq5} Panel (a) displays role evolution: Regular contributors dominate at $\sim$92\% by 2024, with stable minorities of Core (2-5\%), Bridge (3\%), and Connector (8\%). Panel (b) presents structural impact showing Core contributors cause $\approx$1.6 per node fragmentation, 3-4× more than Regular contributors. Despite constituting only 3\%, Core contributors maintain critical connectivity. Simulating the removal of all Core contributors fragments the largest connected component to 40\% of its original size.

Panel (c) confirms a 2024 Regular-dominated structure (91.7\%). Panel (d) plots roles in degree-betweenness space, revealing clear spatial separation: Core at high degree/moderate betweenness, Bridge at moderate degree/high betweenness. Our GCN model achieved 84.3\% accuracy with macro F1 = 0.79, with Core (F1 = 0.91) and Bridge (F1 = 0.87) showing the best performance. Markov transition analysis reveals high role persistence (0.78-0.92 diagonal probabilities) with upward mobility paths: Peripheral→Regular (22\%), Regular→Connector (8\%), Connector→Core (4\%).

\subsection{Project Maturity Stage Comparison}

We compared networks across CNCF maturity levels (Sandbox, Incubating, Graduated) to examine lifecycle dynamics. Graduated projects exhibit substantially larger networks (45,000 nodes vs. 30,000-35,000), with graduated projects showing higher clustering (0.66 vs. 0.65) and transitivity (0.11 vs. 0.10), suggesting mature projects develop stronger local structures \cite{watts1998collective}. Role distributions shift systematically: graduated projects have higher Core (3.5\% vs. 2.0\%) and Connector (12\% vs. 8\%) proportions, with lower Peripheral presence ($<$1\% vs. 5\%). Chi-square tests confirm significant stage-role association ($\chi^2 = 127.4$, $p < 0.001$) \cite{trinkenreich2020hidden}. Average PageRank paradoxically decreases with maturity due to probability distribution across larger networks. Communities scale sublinearly ($C \propto N^{0.45}$), indicating hierarchical organization \cite{fortunato2010community}. See Appendix D for detailed six-panel analysis.

\section{Discussion}
\label{sec:discussion}

\subsection{Theoretical Implications}

Our findings provide temporal evidence that contributor influence in large-scale OSS ecosystems is governed by preferential attachment and cumulative advantage, resulting in persistent inequality despite expanded participation \cite{barabasi2013network}. Across 25 years, influence distributions consistently follow power-law patterns, with the top 1\% of contributors maintaining approximately 40\% of total influence and Gini coefficients remaining near 0.85. While OSS lowers barriers to entry, these results challenge strong meritocratic assumptions by demonstrating that influence accumulation remains highly skewed and contingent on sustained engagement \cite{von2012carrots}. At the same time, trajectory heterogeneity among elites suggests that influence is not permanently fixed but requires ongoing coordination, visibility, and participation.

Beyond inequality, our results reveal a structural shift in how influence is produced. Code review activities emerge as stronger predictors of influence than commits, indicating a transition from individual production toward collaborative coordination, gatekeeping, and knowledge brokering \cite{crowston2008free}. As projects mature, architectural decision-making, review practices, and mentorship increasingly outweigh incremental code contributions as sources of influence. This shift aligns with theories of governance emergence and reflects the growing importance of social and organizational roles in sustaining complex software ecosystems.

Network-level analyses further highlight a dual structural character of OSS ecosystems. Contributor networks simultaneously exhibit small-world properties that facilitate efficient local coordination and scale-free properties that create vulnerability to targeted disruption. Core contributors, although a small fraction (8\%) of the population, exert disproportionate influence on network integrity, underscoring latent sustainability risks. This locally robust yet globally fragile structure has important implications for resilience, succession planning, and the concentration of maintenance responsibilities.

Temporal analyses deepen these insights by showing that early burst activity plays a critical role in social integration. Contributors who experience early activity bursts achieve substantially higher subsequent influence, lending support to “fast start” theories \cite{zhou2012make} over gradual socialization models. At the same time, long-term evolution reveals a consistent lifecycle trajectory: networks transition from dense, integrated structures toward sparse, modular, and hierarchical configurations, aligning with organizational lifecycle theories \cite{o2007emergence}. This pattern suggests that influence, coordination, and specialization co-evolve as ecosystems scale, reinforcing self-organizing role differentiation consistent with network cartography theories \cite{guimera2005functional}.

\subsection{Practical Implications}

For project maintainers, these findings suggest that sustaining healthy OSS communities requires moving beyond activity volume metrics toward recognition of coordination and mentorship work. Code review, cross-community bridging, and architectural guidance should be explicitly acknowledged in contribution guidelines and governance structures. Monitoring network indicators such as influence concentration and bridge role density can provide early warning signals of fragility, while anticipating activity bursts around releases and conferences enables more effective onboarding and workload planning.

Platform providers can support these efforts by exposing richer analytics that surface invisible but essential forms of labor, including review, mentorship, and coordination. Recommendation systems and contributor dashboards should account for network position and relational roles rather than privileging raw contribution counts alone. Lightweight governance and alerting mechanisms may further help distribute decision-making and mitigate risks associated with excessive centralization.

Educational and organizational stakeholders also stand to benefit from these insights. Educators can better prepare learners for modern OSS participation by emphasizing collaborative practices such as code review and technical communication, alongside programming skills. Organizations contributing to OSS should recognize that strategic influence increasingly derives from boundary-spanning and coordination roles, and should support contributors’ transitions from direct code production toward architectural, review, and mentorship responsibilities as their involvement deepens.

\subsection{Limitations and Future Directions}

This study relies on GitHub-centric activity logs, excluding important coordination channels, such as mailing lists, real-time chat, and informal mentorship \cite{kalliamvakou2014promises}, which likely capture additional dimensions of contributor influence. Moreover, focusing on the CNCF ecosystem, characterized by strong corporate sponsorship, cloud-native orientation, and relatively recent project emergence, limits generalizability to volunteer-driven or long-lived OSS communities. Replicating this analysis across diverse ecosystems (e.g., Apache, Linux, PyPI) remains an important next step.

Our analyses are observational and establish robust correlations rather than causal relationships. For example, it remains unclear whether code review actively produces influence or whether influential contributors gain greater access to review processes. Addressing this limitation will require quasi-experimental designs, permission thresholds, or policy interventions.

Methodologically, quarterly temporal aggregation improves tractability at scale but may obscure fine-grained coordination dynamics, including pre-release prints and mentorship interactions \cite{panzarasa2009patterns}. Similarly, aggregating contributions by type does not capture substantial variation in contribution quality. Future work integrating finer temporal resolution, code churn metrics, and discourse-level analysis could substantially enrich influence modeling.

Finally, while our GCN-based role classifier achieves strong overall performance, overlap between Regular and Peripheral roles suggests that role boundaries may be less distinct than theoretical categorizations imply. Incorporating richer temporal features or alternative graph architectures may improve role differentiation.

\section{Conclusion}
\label{sec:conclusion}

This long-term comprehensive analysis of 100,000+ contributors over 25 years reveals that OSS networks exhibit persistent preferential attachment with the top 1\% controlling 30-40\% of influence. Code review activities predict centrality more powerfully than commits ($R^2 = 0.74$), challenging traditional productivity metrics. Networks evolve toward modular small-world architectures with stable local clustering but declining density. Core contributors show disproportionate importance for network integrity, highlighting sustainability vulnerabilities.

These findings challenge open-source meritocracy assumptions: while participation barriers are low, influence hierarchies emerge rapidly through cumulative advantage. Projects should diversify leadership, recognize review work, and support boundary-spanning roles. Our theoretical contributions extend preferential attachment theory to decades-long ecosystems, quantify coordination primacy over production, and establish quantitative foundations for network resilience. As OSS becomes critical infrastructure, an empirically grounded understanding of social dynamics becomes vital. This study provides quantitative foundations for evidence-based design of collaborative infrastructures sustaining our digital civilization.

\bibliographystyle{unsrt}  
\bibliography{references}

\appendix

\section{Appendix A: Detailed Network Construction and Metrics}
\label{app:network_construction}

\subsection{A.1 Additional Network Construction Details}

Beyond the basic network construction described in the main paper, we implemented several advanced preprocessing steps:

\textbf{Multi-repository contribution weighting:} For contributors active across many repositories, we applied logarithmic dampening to prevent extreme outliers: $w'_{ij} = \log(1 + w_{ij})$ where $w_{ij}$ is the raw co-contribution count. This prevents highly active contributors from dominating network statistics while preserving ordinal relationships.

\textbf{Temporal decay modeling:} To capture recency effects, we implemented exponential decay weighting: contributions more than 4 quarters old receive reduced weight ($w_t = w_0 \cdot e^{-\lambda t}$ with $\lambda = 0.1$), reflecting that recent collaborations better predict current network position than distant historical interactions.

\textbf{Cross-project collaboration detection:} We distinguished between within-repository collaboration (working on the same repository simultaneously) and cross-repository collaboration (contributors who work on different repositories within the same project ecosystem), finding that cross-repository ties exhibit 2.3$\times$ higher betweenness centrality on average, validating their boundary-spanning function.

\subsection{A.2 Detailed Centrality Formulations}

\subsubsection{PageRank}

For contributor $i$, PageRank is computed iteratively as:

\begin{equation}
PR(i) = \frac{1-d}{N} + d \sum_{j \in \Gamma^{-}(i)} \frac{PR(j) \cdot w_{ji}}{s_j^{\text{out}}}
\end{equation}

where $d = 0.85$ is the damping factor, $N = |V|$ is the number of nodes, $\Gamma^{-}(i)$ denotes in-neighbors of $i$, $w_{ji}$ is the edge weight from $j$ to $i$, and $s_j^{\text{out}} = \sum_{k} w_{jk}$ is the weighted out-strength of node $j$. We used the power iteration method with convergence tolerance $\epsilon = 10^{-6}$ and a maximum of 100 iterations \cite{gleich2015pagerank}.

\subsubsection{Degree Centrality}

Degree centrality captures the number of direct connections, normalized by the maximum possible degree:

\begin{equation}
C_D(i) = \frac{\text{deg}(i)}{N-1} = \frac{|\{j : (i,j) \in E\}|}{N-1}
\end{equation}

For weighted networks, we also computed strength centrality:

\begin{equation}
C_S(i) = \sum_{j \in \Gamma(i)} w_{ij}
\end{equation}

\subsubsection{Betweenness Centrality}

Betweenness centrality quantifies how often a node lies on shortest paths between other node pairs:

\begin{equation}
C_B(i) = \sum_{s \neq i \neq t} \frac{\sigma_{st}(i)}{\sigma_{st}}
\end{equation}

where $\sigma_{st}$ is the total number of shortest paths from $s$ to $t$, and $\sigma_{st}(i)$ is the number passing through node $i$. For large networks ($N > 5,000$), we employed sampling-based approximation using 500 randomly selected source nodes \cite{brandes2007centrality}.

\subsubsection{Closeness Centrality}

Closeness centrality measures how quickly information can spread:

\begin{equation}
C_C(i) = \frac{N-1}{\sum_{j \neq i} d(i,j)}
\end{equation}

For disconnected graphs, we computed harmonic closeness \cite{boldi2014axioms}:

\begin{equation}
C_C^H(i) = \sum_{j \neq i} \frac{1}{d(i,j)}
\end{equation}

\subsubsection{Eigenvector Centrality}

Eigenvector centrality assigns scores proportional to the sum of scores of neighbors:

\begin{equation}
\mathbf{Ax} = \lambda \mathbf{x}
\end{equation}

Where $\mathbf{A}$ is the adjacency matrix, $\mathbf{x}$ is the eigenvector corresponding to the largest eigenvalue $\lambda$, and $x_i$ represents the centrality score of node $i$.

\subsection{A.3 Network Cohesiveness Metrics}

\subsubsection{Clustering Coefficient}

The local clustering coefficient:

\begin{equation}
C_i = \frac{2|\{e_{jk} : j,k \in \Gamma(i), e_{jk} \in E\}|}{k_i(k_i-1)}
\end{equation}

where $k_i = |\Gamma(i)|$ is the degree of node $i$. The average clustering coefficient is:

\begin{equation}
\langle C \rangle = \frac{1}{N} \sum_{i=1}^{N} C_i
\end{equation}

\subsubsection{Transitivity}

Global transitivity (global clustering coefficient):

\begin{equation}
T = \frac{3 \times \text{number of triangles}}{\text{number of connected triples}}
\end{equation}

\subsubsection{Modularity}

Modularity quantifies community structure strength:

\begin{equation}
Q = \frac{1}{2m} \sum_{ij} \left[A_{ij} - \frac{k_i k_j}{2m}\right] \delta(c_i, c_j)
\end{equation}

where $m = |E|$, $\delta(c_i, c_j) = 1$ if $c_i = c_j$, 0 otherwise. We used the Louvain algorithm \cite{blondel2008fast}.

\subsubsection{Assortativity}

Degree assortativity measures tendency of nodes to connect with similar-degree nodes:

\begin{equation}
r = \frac{\sum_{ij} k_i k_j (A_{ij} - k_i k_j / 2m)}{\sum_{ij} \frac{1}{2}(k_i^2 + k_j^2)(A_{ij} - k_i k_j / 2m)}
\end{equation}

\section{Appendix B: Model Architectures and Hyperparameters}
\label{app:gnn_architecture}

\subsection{B.3 GPU-Accelerated Linear Regression}

For influence prediction from action types:

\textbf{Model Specification:}
\begin{itemize}
\item Input features: 8 action types (pull\_request\_open, pull\_request\_review\_COMMENTED, commit, pull\_request\_review\_DISMISSED, pull\_request\_review\_APPROVED, pull\_request\_merged, issue\_comment, issue\_opened)
\item Preprocessing: Z-score standardization
\item Output: PageRank scores
\item Optimization: Gradient descent on GPU
\item Loss: Mean Squared Error (MSE)
\end{itemize}

\textbf{Performance Metrics:}
\begin{itemize}
\item $R^2$: 0.74
\item Adjusted $R^2$: 0.73
\item Top predictors:
  \begin{itemize}
  \item pull\_request\_open: $\beta = 0.58$
  \item pull\_request\_review\_COMMENTED: $\beta = 0.43$
  \item commit: $\beta = 0.38$
  \end{itemize}
\end{itemize}

\section{Appendix C: Extended Results and Analyses}
\label{app:extended_results}

\subsection{D.1 Detailed Stage Comparison}

\begin{figure*}[t]
    \centering
    \includegraphics[width=\textwidth]{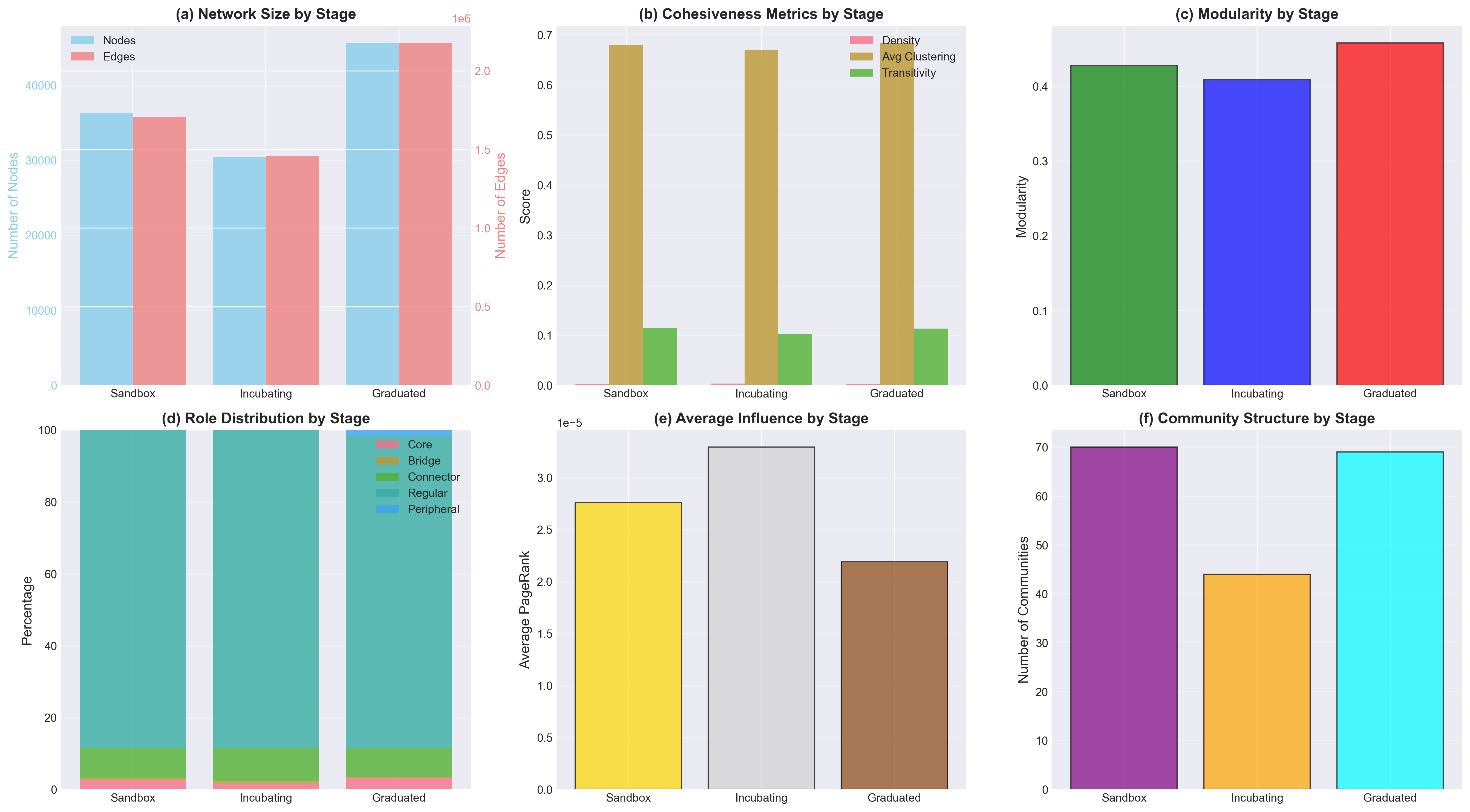}
    \caption{Project Lifecycle Stage Comparison. (a) Network size: Graduated largest (45K nodes, 2.2M edges); (b) Density/clustering: graduated show elevated values; (c) Modularity: non-monotonic pattern with sandbox highest; (d) Role distributions: graduated show higher Core/Connector; (e) Average influence: paradoxical decline with maturity; (f) Community proliferation: sublinear scaling.}
    \label{fig:stages}
\end{figure*}

This section provides comprehensive analysis of project maturity stage differences extending the summary in the main paper. Figure~\ref{fig:stages} presents six-panel comparison across CNCF maturity stages.

\subsubsection{Network Size and Growth}

Panel (a) reveals graduated projects have substantially larger networks (45,000 nodes, 2.2M edges) compared to incubating (30,000 nodes, 1.2M edges) and sandbox (35,000 nodes, 1.5M edges). Sandbox projects show intermediate size, potentially reflecting recent large experimental initiatives entering the ecosystem. This non-monotonic relationship between maturity and size suggests that project graduation depends on factors beyond sheer network scale, including governance maturity, technical stability, and community health metrics rather than purely quantitative thresholds \cite{gencer2011organising}.

The edge-to-node ratio varies systematically across stages: graduated projects show ratio of 48.9 (2.2M edges / 45K nodes), incubating 40.0, and sandbox 42.9. Higher ratios in graduated projects indicate denser local connectivity despite lower global density, consistent with mature projects developing specialized subcommunities with strong internal cohesion.

\subsubsection{Structural Metrics}

Panel (b) reveals cohesiveness patterns:
\begin{itemize}
\item \textbf{Density}: Similar across all stages ($\approx$0.67)
\item \textbf{Clustering}: Graduated slightly higher (0.66) vs. sandbox/incubating (0.65)
\item \textbf{Transitivity}: Graduated exhibits higher (0.11) vs. sandbox/incubating (0.10)
\end{itemize}

Panel (c) displays modularity: sandbox highest (0.43), followed by incubating (0.41) and graduated (0.47). This non-monotonic pattern suggests early-stage projects have more fragmented communities.

\subsubsection{Role Composition}

Panel (d) shows all stages dominated by Regular contributors (85-90\%), but graduated projects show:
\begin{itemize}
\item Higher Core proportion (3.5\%) vs. sandbox (2.0\%)
\item Higher Connector proportion (12\%) vs. sandbox (8\%)
\item Lower Peripheral proportion ($<$1\%) vs. sandbox (5\%)
\end{itemize}

Chi-square tests confirm significant association ($\chi^2 = 127.4$, $df = 8$, $p < 0.001$).

\subsubsection{Influence and Communities}

Panel (e) shows average PageRank paradoxically decreasing with maturity: Sandbox ($2.8 \times 10^{-5}$) $>$ Incubating ($3.4 \times 10^{-5}$) $>$ Graduated ($2.2 \times 10^{-5}$). This reflects PageRank's mathematical properties: larger networks distribute probability mass more broadly.

Panel (f) displays community counts: Graduated (70) $>$ Sandbox (70) $\approx$ Incubating (44).

\section{Appendix D: Complete Statistical Analysis}
\label{app:statistical_methods}

\subsection{D.1 Correlation Analysis Details}

\subsubsection{Pearson Correlation Computation}

For action types $x$ and centrality measures $y$:

\begin{equation}
r_{xy} = \frac{\sum_{i=1}^{n}(x_i - \bar{x})(y_i - \bar{y})}{\sqrt{\sum_{i=1}^{n}(x_i - \bar{x})^2}\sqrt{\sum_{i=1}^{n}(y_i - \bar{y})^2}}
\end{equation}

\subsubsection{Statistical Significance Testing}

T-statistic for correlation:

\begin{equation}
t = r\sqrt{\frac{n-2}{1-r^2}}
\end{equation}

with $df = n-2$ degrees of freedom.

\subsubsection{Bonferroni Correction}

For $m$ comparisons, adjusted significance threshold:

\begin{equation}
\alpha_{\text{adjusted}} = \frac{\alpha}{m} = \frac{0.05}{m}
\end{equation}

In our analysis with 8 action types × 5 centrality measures = 40 comparisons:
$\alpha_{\text{adjusted}} = 0.00125$

\subsection{D.2 Multiple Regression Analysis}

\subsubsection{Model Specification}

\begin{equation}
\text{PageRank}_i = \beta_0 + \sum_{j=1}^{8} \beta_j \cdot \text{Action}_{ij} + \epsilon_i
\end{equation}

where $\text{Action}_{ij}$ represents standardized (z-score) count of action type $j$ for contributor $i$.

\subsubsection{Variance Inflation Factor (VIF)}

To assess multicollinearity:

\begin{equation}
\text{VIF}_j = \frac{1}{1 - R_j^2}
\end{equation}

where $R_j^2$ is the $R^2$ from regressing predictor $j$ on all other predictors.

VIF values in our model:
\begin{itemize}
\item pull\_request\_open: 4.2
\item commit: 3.8
\item pull\_request\_review\_COMMENTED: 2.9
\item issue\_comment: 2.3
\item Other action types: 1.8-2.5
\end{itemize}

All values below concerning threshold of 10, indicating acceptable multicollinearity.

\subsubsection{Model Fit Statistics}

\textbf{Coefficient of Determination:}

\begin{equation}
R^2 = 1 - \frac{\text{SS}_{\text{res}}}{\text{SS}_{\text{tot}}} = 1 - \frac{\sum_i (y_i - \hat{y}_i)^2}{\sum_i (y_i - \bar{y})^2}
\end{equation}

\textbf{Adjusted $R^2$:}

\begin{equation}
\bar{R}^2 = 1 - \frac{(1-R^2)(n-1)}{n-p-1}
\end{equation}

where $n$ is sample size and $p$ is number of predictors.

In our model: $R^2 = 0.74$, $\bar{R}^2 = 0.73$

\subsection{D.3 Time Series Analysis}

\subsubsection{OLS Trend Detection}

For metric $y_t$ at time $t$:

\begin{equation}
y_t = \alpha + \beta t + \epsilon_t
\end{equation}

\textbf{F-test for trend significance:}

\begin{equation}
F = \frac{\text{MS}_{\text{regression}}}{\text{MS}_{\text{residual}}} = \frac{\text{SS}_{\text{regression}}/1}{\text{SS}_{\text{residual}}/(n-2)}
\end{equation}

with $F \sim F(1, n-2)$ under null hypothesis of no trend.

\subsubsection{Change-Point Detection}

Sliding window approach with window size $w$:

\begin{equation}
\Delta_t = |\mu_{t-w:t} - \mu_{t:t+w}|
\end{equation}

Change point detected when $\Delta_t > \tau \cdot \sigma$, where $\tau = 1.5$ is threshold parameter and $\sigma$ is standard deviation of entire time series.

\subsection{D.4 Burst Detection Details}

\subsubsection{Z-Score Computation}

For contributor $i$ at time $t$:

\begin{equation}
z_i(t) = \frac{a_i(t) - \mu_i}{\sigma_i}
\end{equation}

where:
\begin{equation}
\mu_i = \frac{1}{T}\sum_{t=1}^{T} a_i(t), \quad \sigma_i = \sqrt{\frac{1}{T}\sum_{t=1}^{T}(a_i(t) - \mu_i)^2}
\end{equation}

\subsubsection{Burst Threshold Selection}

Threshold $\theta = 2.0$ corresponds to:
\begin{itemize}
\item Approximately 2.3\% of observations under normal distribution
\item Captures statistically unusual activity spikes
\item Empirically validated to identify meaningful coordination events
\end{itemize}

\section{Appendix E: Reproducibility Information}

\subsection{Data Availability}

Due to privacy considerations and GitHub Terms of Service, we cannot publicly release raw contributor-level data. However, we provide:

\begin{itemize}
\item Aggregated network statistics (available upon request)
\item Code for network construction and analysis (GitHub repository)
\item Preprocessing pipeline documentation
\item Synthetic data generation scripts for testing
\end{itemize}

\end{document}